\documentclass[runningheads]{llncs}

\usepackage{eccv}

\usepackage{eccvabbrv}

\usepackage{graphicx}
\usepackage{booktabs}
\usepackage{bbm}
\usepackage{makecell}
\usepackage{multirow}
\usepackage{subcaption}
\usepackage{microtype}
\usepackage{algorithm}
\usepackage{algpseudocode} % For algorithmicx style pseudocode
\usepackage[accsupp]{axessibility}  % Improves PDF readability for those with disabilities.

\usepackage{hyperref}

\usepackage{orcidlink}

\begin{document}


\title{Learning the Unlearned: Mitigating Feature Suppression in Contrastive Learning} 

\titlerunning{Learning the Unlearned}

\renewcommand{\thefootnote}{\relax}
\footnotetext[0]{* Equal Contribution~ $\dagger$ Correspondence to: Mengling Feng <\email{ephfm@nus.edu.sg}>, Bryan Hooi <\email{bhooi@comp.nus.edu.sg}>.~ Xiang Lan and Mengling Feng are affiliated with Saw Swee Hock School of Public Health and Institute of Data Science, NUS.~ Part of this work was done during Jihai Zhang's internship at Shanghai AI Laboratory.}
\author{Jihai Zhang \inst{*,1},
Xiang Lan \inst{*,2},
Xiaoye Qu\inst{3},
Yu Cheng \inst{1},\\
Mengling Feng \inst{2,\dagger},
Bryan Hooi \inst{2,\dagger}}

\renewcommand{\thefootnote}{\arabic{footnote}}

\authorrunning{J.~Zhang, X.~Lan, et al.}

\institute{
$^1$The Chinese University of Hong Kong\\ $^2$National University of Singapore $^3$Shanghai AI Laboratory
}

\maketitle
\begin{abstract}
  Self-Supervised Contrastive Learning has proven effective in deriving high-quality representations from unlabeled data. 
  However, a major challenge that hinders both unimodal and multimodal contrastive learning is feature suppression, a phenomenon where the trained model captures only a limited portion of the information from the input data while overlooking other potentially valuable content.
  This issue often leads to indistinguishable representations for visually similar but semantically different inputs, adversely affecting downstream task performance, particularly those requiring rigorous semantic comprehension.
  To address this challenge, we propose a novel model-agnostic \textbf{M}ultistage \textbf{C}ontrastive \textbf{L}earning (MCL) framework. 
  Unlike standard contrastive learning which inherently captures one single biased feature distribution, MCL progressively learns previously unlearned features through \textit{feature-aware negative sampling} at each stage, where the negative samples of an anchor are exclusively selected from the cluster it was assigned to in preceding stages. Meanwhile, MCL preserves the previously well-learned features by \textit{cross-stage representation integration}, integrating features across all stages to form final representations.
  Our comprehensive evaluation demonstrates MCL's effectiveness and superiority across both \textbf{unimodal} and \textbf{multimodal} contrastive learning, spanning a range of model architectures from ResNet to Vision Transformers (ViT).
  Remarkably, in tasks where the original CLIP model has shown limitations, MCL dramatically enhances performance, with improvements up to threefold on specific attributes in the recently proposed MMVP benchmark. Codes are available at \url{https://github.com/MajorDavidZhang/MCL.git}.
\end{abstract}

\section{Introduction}
\begin{figure}[!ht]
    \centering
    \begin{subfigure}{0.45\textwidth}
        \centering  
        \includegraphics[width=0.9\textwidth]{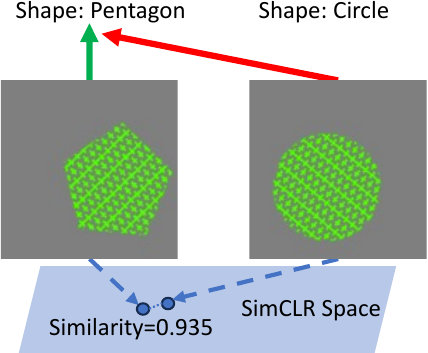}
        \caption{Images from Trifeature with different shapes and textures have high similarity in the SimCLR space.}
        \label{fig:fs_simclr}
    \end{subfigure}
    \hfill
    \begin{subfigure}{0.45\textwidth}
        \centering
        \includegraphics[width=0.9\textwidth]{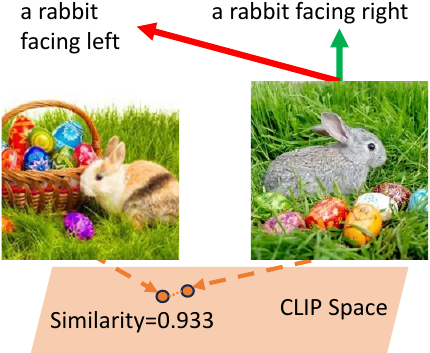}
        \caption{Images from MMVP with different orientations and directions have high similarity in the CLIP space.}
        \label{fig:fs_clip}
    \end{subfigure}
    \caption{Demonstration of feature suppression in both unimodal (SimCLR) and multimodal (CLIP) settings. The green arrows refer to correct linear evaluation classification/ pairing; the red arrows refer to incorrect ones.}
    \label{fig:fs}
\end{figure}

Self-Supervised contrastive learning obtains high-quality representations by maximizing the similarity between an anchor and its associated positive samples, while concurrently increasing the separation among the dissimilar data samples in the embedding space~\cite{jaiswal2020survey}. Various contrastive learning models serve as fundamental pretrained backbones across different fields~\cite{radford2021learning,chen2020simple,he2020momentum,woo2022cost}. 
However, recent studies~\cite{robinson2021can,tamkin2022feature, xue2023features} have shown that representations derived from standard contrastive learning often miss substantial portions of input information. This phenomenon is referred to as feature suppression.
Such suppression can severely compromise the effectiveness of models in various downstream tasks, ranging from classification~\cite{robinson2021can} to pattern recognition~\cite{tong2024eyes}. 
% 
% Furthermore, as multimodal large language models (MLLMs) predominantly employ the CLIP~\cite{radford2021learning} model, trained via contrastive learning, as their vision encoder~\cite{liu2023visual,zhu2023minigpt,yin2023survey}, the analogous feature suppression issue in CLIP hampers these MLLMs’ ability to differentiate between images with varying semantics, leading to severe hallucination problem in MLLMs~\cite{tong2024eyes}.
In addition, the feature suppression issues are also observed in multimodal contrastive learning~\cite{bleeker2023reducing, tong2024eyes}, such as CLIP~\cite{radford2021learning}, which is predominantly adopted in current multimodal large language models(MLLMs)~\cite{liu2023visual,zhu2023minigpt,yin2023survey} as the vision encoder. Feature suppression in CLIP significantly impedes the capability of MLLMs to differentiate between images with varying semantics, resulting in severe hallucination problems~\cite{tong2024eyes,li2023evaluating,liu2023mitigating,leng2023mitigating} within these models.

Referencing \cref{fig:fs_simclr}, SimCLR trained on the Trifeature dataset fails to differentiate between a circle and a pentagon with the same shape and color. Similarly, the OpenAI pretrained CLIP model struggles to distinguish between a rabbit facing left and a rabbit facing right due to the high similarity of their representations in the embedding space, as illustrated in \cref{fig:fs_clip}. Consequently, MLLMs that utilize CLIP as their vision encoder experience systematic failures on tasks involving such distinctions.
% 
% Although feature suppression is a critical issue in contrastive learning, only a handful of methods have been proposed to address it, often at the expense of compromising the integrity of previously well-learned features~\cite{robinson2021can,tamkin2022feature} or by incorporating an additional reconstruction loss~\cite{bleeker2023reducing} which is not feasible for large-scale settings such as CLIP due to its computational demands. 
While feature suppression presents a critical challenge in contrastive learning, there are only a handful of methods proposed to address it. These approaches often come at the expense of compromising previously well-learned features~\cite{robinson2021can,tamkin2022feature}. Alternatively, they necessitate an additional reconstruction loss~\cite{bleeker2023reducing}, which is rendered impractical for large-scale applications such as CLIP due to the high computational demands.
% Moreover, existing solutions typically focus solely on either unimodal or multimodal contrastive learning without general consideration.
Furthermore, the scope of these existing methodologies is often restricted to either unimodal or multimodal contrastive learning, lacking universal applicability.

In this paper, we propose a novel model-agnostic framework: Multistage Contrastive Learning (MCL), designed to effectively tackle feature suppression in both \textbf{unimodal} and \textbf{multimodal} settings.
Unlike standard single-stage contrastive learning that often collapses to certain features, 
MCL aims to progressively learn new features that have not been explored in the previous training stages, while retaining the well-learned features. 
Throughout the multistage training process, we implement a \textit{feature-aware negative sampling} strategy designed to compel the model towards exploring unlearned features in earlier stages. 
Inspired by the observation that representations in contrastive learning tend to cluster according to the learned features~\cite{robinson2021can,radford2021learning}, at each stage, MCL selects negative samples for each anchor exclusively from the cluster it was assigned to in preceding stages. Because samples within the same cluster share similar learned features, these features cannot be reused to accomplish the contrastive learning objective: to discriminate the anchor from negative samples. Therefore, the model is necessitated to discover and incorporate previously unlearned features to fulfill the contrastive learning objective.
After the multistage learning process, \textit{cross-stage representation integration} is employed. Here the representations of data samples from all stages are concatenated to form the ultimate representations, ensuring that the well-learned features are retained.
%The cluster assignments are determined by performing K-means on the representations encoded by the trained model in the preceding stage. 
% 

In summary, the contributions of this work are three-fold. 
\textit{First}, we propose a novel model-agnostic contrastive learning framework: Multistage Contrastive Learning that mitigates the severe issues of feature suppression commonly encountered in contrastive learning.
\textit{Second}, to the best of our knowledge, this is the first work to discuss and address the problem of feature suppression in both unimodal and multimodal contrastive learning. 
\textit{Third}, we empirically demonstrate that the proposed MCL can be adapted to various contrastive learning settings and further boost their performance using different encoder backbones scaling from ResNet-18 to ViT-L-14. Notably, MCL demonstrates a significant improvement and boosts the average accuracy from $20.0$ to $32.6$ in the CLIP setting on the MMVP benchmark.

\section{Related Works}
\subsection{Contrastive Learning}

How to extract useful information from unlabeled data is an important question in machine learning~\cite{bengio2013representation}. Among all branches of methods, contrastive learning flourishes in recent years~\cite{jaiswal2020survey} and plays an important role in text-to-image generation~\cite{radford2021learning,ramesh2021zero,ramesh2022hierarchical} and multimodal large language models~\cite{liu2023visual,zhu2023minigpt,yin2023survey}. Contrastive learning aims to learn useful representations from unlabeled data by maximizing the agreement between different views of the data~\cite{federici2020learning}. 
Recently, different variants of contrastive learning methods have been proposed~\cite{chen2020simple, he2020momentum,oord2018representation,caron2020unsupervised}. 
CPC~\cite{oord2018representation} learns representations by predicting future samples in a sequence using an auto-regressive model with a contrastive loss.
MoCo~\cite{he2020momentum} is designed to overcome the limitations of batch size in contrastive learning by introducing a dynamic dictionary with a queue and a moving-averaged encoder. 
SimCLR~\cite{chen2020simple} is a simple yet effective framework for contrastive learning of visual representations. SimCLR demonstrated that, with sufficiently large batch sizes, it is possible to learn powerful representations without needing specialized architectures or a memory bank.
MoCo-v2~\cite{chen2020improved} builds upon the original MoCo by incorporating several improvements from SimCLR. 
Negative-free contrastive learning~\cite{chen2021exploring,grill2020bootstrap} further simplifies contrastive learning by removing the requirement of explicit negative samples. Although contrastive learning achieves promising performance in many fields~\cite{van2021unsupervised, yue2022ts2vec, shvetsova2022everything,woo2022cost,Lan_Ng_Hong_Feng_2022,lan2024towards}, it cannot guarantee all semantically relevant features are learned when multiple features exist~\cite{chen2021intriguing, xiao2020should, xue2023features,bleeker2023reducing,tong2024eyes}. 

\subsection{Feature Suppression in Contrastive Learning}
% When multiple features exist, the contrastive learning encoder may only encode the information of some features while ignoring the other features. 
The feature suppression phenomenon has been first empirically observed by Chen \etal~\cite{chen2021intriguing}. Subsequently, Robinson \etal~\cite{robinson2021can} formally brings up this problem as feature suppression, and shows that simply minimizing the InfoNCE loss cannot avoid feature suppression. Both Robinson \etal~\cite{robinson2021can} and Kukleva \etal~\cite{kukleva2023temperature} observed that the temperature parameter affects the trade-off of which features are learned and which features are suppressed. Xiao \etal~\cite{xiao2020should} have discovered that certain augmentations used to generate positive samples might destroy the feature information, hence hindering the learning process of corresponding features. Assran \etal~\cite{assran2022hidden} point out that feature suppression might be caused by the hidden prior distribution bias in contrastive learning. Xue \etal~\cite{xue2023features} demonstrate that the simplicity bias of stochastic gradient descent is one of the factors. Not only in the above unimodal setting, Bleeker \etal~\cite{bleeker2023reducing} first studies this problem in the multimodal setting. Tong \etal~\cite{tong2024eyes} have observed that images with different semantics have unreasonably high similarities in CLIP~\cite{radford2021learning} embedding space, which is also highly related to feature suppression in multimodal contrastive learning. 

Few methods specifically address the challenge of feature suppression in contrastive learning. 
Robinson \etal~\cite{robinson2021can} proposed a technique aimed at eliminating whichever features distinguish the positive sample from negative samples in the embedding space. Similarly, Tamkin \etal~\cite{tamkin2022feature} applied a comparable strategy but targeted the input space. Both approaches are based on adversarial training, which does not ensure the preservation of previously well-learned features. 
% They also require training models from scratch—a process notably resource-intensive, especially in large-scale applications like CLIP~\cite{radford2021learning}. 
In contrast, our approach diverges by sequentially learning new features stage by stage without compromising the integrity of already learned features. 
% Moreover, we show that in large-scale environments, fine-tuning existing models through our multistage framework not only yields effective results but also significantly reduces computational demands.
%
Bleeker \etal~\cite{bleeker2023reducing} mitigate the feature suppression problem by introducing an additional reconstruction loss, which is not feasible for large-scale settings such as CLIP due to the high computational cost. In contrast, our approach does not necessitate modifications to the original loss function or alterations to the base model, conserving computational resources and ensuring compatibility across different models.
It is worth noting that, unlike the aforementioned methods and several other related approaches~\cite{mishra2020learning, chen2021incremental, chu2023image, ge2021robust, shah2022max} that are confined to either unimodal or multimodal settings, our work stands as the first attempt to tackle feature suppression across both unimodal and multimodal contrastive learning.

\section{Preliminaries}
\subsection{Self-Supervised Contrastive Learning}
In contrastive learning, the core objective is to minimize the distance between positive pairs while maximizing the distance between negative pairs within the representation space. This objective compels the model to effectively distinguish positive pairs from their negative counterparts. Without loss of generality, here we consider the basic yet effective Noise Contrastive Estimation (NCE)~\cite{gutmann2010noise} based contrastive learning model~\cite{liu2021self} as our backbone contrastive learning model. Given an anchor $\mathbf{x}$, its positive sample  $\mathbf{x}^+$ and $m$ negative samples $\{\mathbf{x}_i^-\}_{i=1}^m$, the model required to minimize the InfoNCE loss defined below:

\begin{align}\label{eq:infornce}
    \mathcal{L} =\mathbb{E}_{\mathbf{x},\mathbf{x}^+,\{\mathbf{x}_i^-\}_{i=1}^m} \left[-\log \frac{e^{s(\mathbf{z},\mathbf{z}^+)/\tau}}
    {e^{s(\mathbf{z},\mathbf{z}^+)/\tau}
    +\sum_{i=1}^m e^{s(\mathbf{z},\mathbf{z}^-_i)/\tau}}
    \right],
\end{align}
where $s(\cdot,\cdot)$ denotes the cosine similarity, $\tau$ represents the temperature, and $\mathbf{z}$, $\mathbf{z}^+$, and $\mathbf{z}^-$ are the corresponding embeddings of $\mathbf{x}$, $\mathbf{x}^+$, and $\mathbf{x}^-$. In unimodal contrastive learning, typically the positive samples are augmentations of the anchor, while in multimodal contrastive learning, the positive samples are usually data pairs with similar semantics as the anchor. The negative samples are randomly sampled data. 

\begin{table}[ht]\footnotesize
    \centering
    \caption{Demonstration of feature suppression in unimodal and multimodal settings.}
    \begin{subtable}{.5\textwidth}
        \centering
        \caption{Linear evaluation results of SimCLR and MoCo-v2 trained on Trifeature and CIFAR-MNIST (C-M). The feature suppression problem in SimCLR is severe. 
        In Trifeature, SimCLR significantly ignores the shape information, and in CIFAR-MNIST, it almost completely neglects the CIFAR information.
        % The shape information in Trifeature is largely ignored by SimCLR, and the CIFAR information in CIFAR-MNIST is totally ignored by SimCLR. 
        % 
        With MoCo-v2, the issue of feature suppression is less pronounced but still exists, considering the two datasets are very simple.}
        \begin{tabular}{l|c c  }
        \toprule
        Feature &  SimCLR  & MoCo-v2   \\
        \midrule 
         C-M(CIFAR) & 0.10 &   0.77\\
         C-M(MNIST) & 0.99 &   0.98 \\
         Trifeature(Shape)   & 0.44 &   0.85   \\
         Trifeature(Texture) & 0.92 &   0.99\\
         Trifeature(Color)   & 1.00 &   1.00\\
        \bottomrule  
        \end{tabular}
        \label{tab:fs_unimodal}
    \end{subtable}
    \hfill
    \begin{subtable}{.47\textwidth}
        \centering
        \caption{Performance of CLIP on the MMVP benchmark. The performance is low on most of the attributes. \textbf{O\&D}: Orientation and Direction, \textbf{PSF}: Presence of Specific Features, \textbf{S\&C}: State and Condition, \textbf{Q\&C}: Quantity and Count, \textbf{P\&R}: Positional and Relational Context, \textbf{C\&A}: Color and Appearance, \textbf{S\&P}: Structural and Physical Characteristics, \textbf{Texts}: Texts, \textbf{V\&P}: Viewpoint and Perspective.}
        \begin{tabular}{l|c|l|c}
        \toprule
        Attribute &  Accuracy  & Attribute &  Accuracy   \\
        \midrule 
         \textbf{O\&D}   &   26.7    &   \textbf{C\&A}  & 40.0  \\
         \textbf{PSF}    &   13.3    &   \textbf{S\&P}  & 26.7    \\
         \textbf{S\&C}   &   26.7    &   \textbf{Texts}   &   13.3    \\
         \textbf{Q\&C}   &   6.7     &   \textbf{V\&P}    &   20.0    \\
         \textbf{P\&R}   &   6.7     &   Average &   20.0    \\
        \bottomrule  
        \end{tabular}
        \label{tab:fs_multimodal}
    \end{subtable}
\end{table}

\subsection{Feature Suppression}
% Feature suppression refers to the phenomenon that, in the presence of multiple features, a contrastive learning encoder selectively captures information from certain features while disregarding others. 
This section empirically illustrates the feature suppression phenomenon across both unimodal and multimodal contrastive learning settings.

In the unimodal setting, the ability of the encoder to capture specific feature information can be assessed through the linear evaluation accuracy of discriminating that feature. A high linear evaluation accuracy for a given feature suggests the encoder has successfully captured substantial information regarding that feature, and vice versa. We train ResNet-18 encoders using SimCLR~\cite{chen2020simple} and MoCo-v2~\cite{chen2020improved} on two datasets (CIFAR-MNIST and Trifeature)\footnote[1]{More detailed dataset description and settings can be found in \cref{sec:experiments}.} following Robinson \etal~\cite{robinson2021can} and Chen \etal~\cite{chen2021intriguing}, to demonstrate the feature suppression phenomenon. The linear evaluation results for each feature across both datasets are shown in \cref{tab:fs_unimodal}. 
In CIFAR-MNIST, the linear evaluation accuracy for MNIST features is high but both SimCLR and MoCo-v2 show comparatively lower accuracy for CIFAR features. This discrepancy indicates MNIST features’ predominance, with CIFAR information being notably overlooked—SimCLR, in particular, demonstrates this trend more pronouncedly.
In Trifeature, a similar pattern emerges: both contrastive learning methods achieve high linear evaluation accuracy for texture and color, yet falter when it comes to shape. This divergence suggests that the encoders while the encoders sufficiently capture texture and color, they neglect shape information. Consequently, as illustrated in \cref{fig:fs_simclr}, images with different shapes in Trifeature have high similarity in the SimCLR embedding space. This overlap significantly hampers the model’s capacity to discern shapes, adversely affecting performance on downstream tasks reliant on shape differentiation.

Similarly, in the multimodal setting, our observations align with those as reported by Tong \etal~\cite{tong2024eyes}. Referencing \cref{fig:fs_clip}, we find that images, despite varying significantly in object orientation and direction, are represented with striking similarity in the CLIP embedding space. As a result, CLIP struggles to discern differences in object orientation and direction. This limitation can lead to orientation-based hallucinations in multimodal large language models that utilize CLIP as their vision encoder~\cite{tong2024eyes}. To evaluate this phenomenon, Tong \etal~\cite{tong2024eyes} introduce the MMVP benchmark.
% which comprises image pairs that have distinctly different semantics but display high similarity in the CLIP space. This benchmark challenges the contrastive learning model to accurately match these challenging image pairs with their corresponding textual descriptions, as exemplified in \cref{fig:fs_clip}. 
We evaluate the OpenAI ViT-L-14 CLIP model~\cite{radford2021learning} with $224^2$ resolution on the MMVP benchmark. 
The results, presented in \cref{tab:fs_multimodal}, show limited performance across a range of attributes, highlighting a severe feature suppression issue of the current CLIP model.
% indicating suboptimal performance across various attributes, 

\section{Multistage Contrastive Learning}
\begin{figure}[ht]
    \centering
    \includegraphics[width=\textwidth]{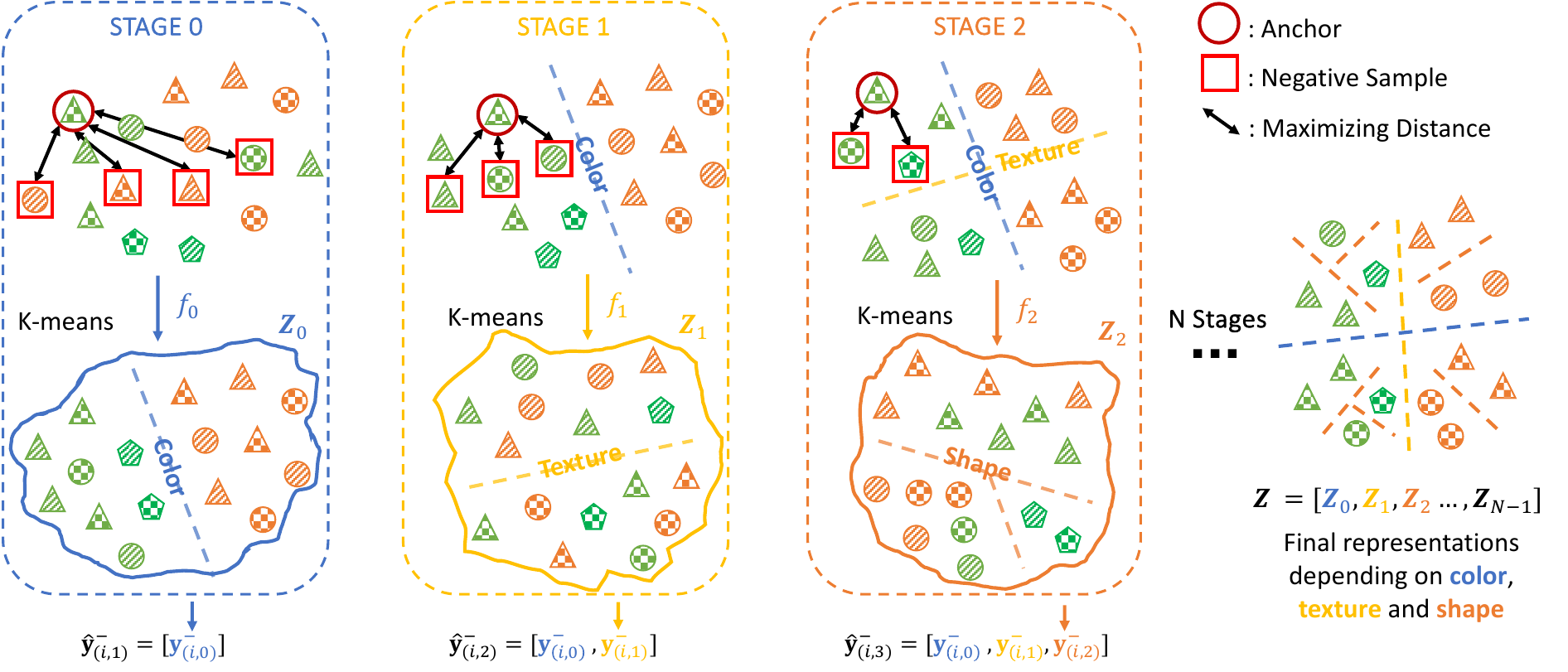}
    \caption{Overview of the Multistage Contrastive Learning (MCL) Framework: Initially, the model is trained and its output representations are clustered. At each subsequent stage, cluster assignments from the previous stages are concatenated to derive a pseudo label. Throughout the training phase, negative samples are selected based on their matching pseudo label with the anchor. The final representations are the concatenation of representations in each stage. For simplicity, only three stages are shown here. The `color', `shape', and `texture' here are used metaphorically to represent abstract features.}
    \label{fig:method}
\end{figure}

\subsection{Feature-aware Negative Sampling}
In the initial phase of MCL, we train an encoder $f_0$ using the standard NCE objective as in Eq.\ref{eq:infornce}. For clarity, we specify that the subscript’s first component refers to the sample index, while the latter signifies the stage index. Considering a dataset with $M$ samples $\boldsymbol{X}=\{\mathbf{x}_i\}_{i=1}^M$, after training we obtain the dataset's encoded representations $\boldsymbol{Z}_0 =\{f_0(\mathbf{x}_i)\}_{i=1}^M $. Subsequently, we apply K-means clustering to $\boldsymbol{Z}_0$ and obtain the initial cluster assignments $\boldsymbol{Y}_0 = \{\mathbf{y}_{(i,0)}\}_{i=1}^M$. The MCL training process consists of $N$ stages. At  $j^{th}$ stage, for a given sample $\mathbf{x}_i$, we define its pseudo label $\hat{\mathbf{y}}_{(i,j)}$ as the concatenation of its cluster assignments from all preceding stages, as follows: 
\begin{equation}\label{eq:cluster}
    \hat{\mathbf{y}}_{(i,j)}=[\mathbf{y}_{(i,0)},\cdots,\mathbf{y}_{(i,j-1)}].
\end{equation}

When feature suppression happens, representations will be clustered by dominant features, since inputs with the same dominant features exhibit high similarity in representation space~\cite{robinson2021can,tong2024eyes}. Therefore, data samples with identical pseudo labels indicate they have similar dominant features learned in prior stages, as illustrated in Fig. \ref{fig:method}.
During the $j^{th}$ stage, the encoder $f_j$ is trained under a refined InfoNCE objective incorporating feature-aware negative sampling: the negative samples must have identical pseudo labels with the anchor. 
Formally, we define the optimization objective for $f_j$ at $j^{th}$ stage as:
\begin{align}\label{eq:infornce_multistage}
    \mathcal{L} =\mathbb{E}_{\mathbf{x},\mathbf{x}^+,\{\mathbf{x}_i^-\}_{i=1}^m}
    \left[-\log \frac{e^{s(\mathbf{z},\mathbf{z}^+)/\tau}}
    {e^{s(\mathbf{z},\mathbf{z}^+)/\tau}
    +\sum_{i=1}^m \mathbbm{1}_{\hat{\mathbf{y}}_j=\hat{\mathbf{y}}_{(i,j)}^-} e^{s(\mathbf{z},\mathbf{z}^-_i)/\tau}}
    \right],
\end{align} 
where $\hat{\mathbf{y}}_j=[\mathbf{y}_{0},\cdots,\mathbf{y}_{j-1}]$ represents the pseudo label of the anchor $\mathbf{x}$ at $j^{th}$ stage; $\hat{\mathbf{y}}_{(i,j)}^-=[\mathbf{y}_{(i,0)}^-,\cdots,\mathbf{y}_{(i,j-1)}^-]$ 
denotes the pseudo label of the negative sample $\mathbf{x}_i^-$ at $j^{th}$ stage; and $\mathbbm{1}_{\hat{\mathbf{y}}_j=\hat{\mathbf{y}}_{(i,j)}^-}\in \{0,1\}$ corresponds to an indicator function evaluating to $1$ if and only if $\hat{\mathbf{y}}_j=\hat{\mathbf{y}}_{(i,j)}^-$. 
Feature-aware negative sampling ensures that the previously dominant features can not be re-utilized by the model to optimize the InfoNCE loss since the anchor and its negative samples share similar dominant features learned in earlier stages. 
As such, the model has to identify and utilize features distinct from those previously dominant features. For example, if an encoder $f$ initially learns color as a dominant feature in Trifeature, through feature-aware negative sampling, the subsequent encoder $f'$ will be tasked with differentiating samples with the same color. This forces the contrastive model to explore alternative features such as texture or shape, rather than relying solely on color, as depicted in \cref{fig:method}.
Following the training of the $j^{th}$ stage, we obtain the representations $\boldsymbol{Z}_j =\{f_j(\mathbf{x}_i)\}_{i=1}^M $ encoded by $f_j$, and the cluster assignments $\boldsymbol{Y}_j = \{\mathbf{y}_{(i,j)}\}_{i=1}^M$ for the feature-aware negative sampling in the next stage.

Notice that derived from Eq.~\ref{eq:cluster}, $N$ stages of clustering with $K$ clusters each lead to a potential total of $K^N$ unique clusters. To ensure a meaningful clustering where clusters are approximately balanced and contain a sufficient number of samples, there's a mathematical constraint on the values of $N$ and $K$:
\begin{equation}\label{eq:kn_bound}
    K^N \leq \frac{M}{b},
\end{equation}
where $M$ represents the total number of samples in the training dataset, and $b$ denotes the batch size. 
% This constraint ensures that the resultant clusters each contain enough samples to form a batch, preventing a cluster from having fewer samples than the batch size and thus becoming impractical for training.
This constraint guarantees that the resulting clusters each have a sufficient number of samples to form a batch, avoiding situations where a cluster contains fewer samples than the batch size, which would make it impractical for training purposes.

\subsection{Cross-stage Representation Integration}
Upon completing the training across all stages, we employ cross-stage representation integration to derive the final representation, which aims to preserve the information of well-learned features from each stage. Specifically, we element-wise concatenate the representations encoded by each trained encoder, resulting in comprehensive final representations for downstream tasks. The cross-stage representation integration is defined as:
\begin{equation}
    \boldsymbol{Z} = \{[f_0(\mathbf{x}_i),\cdots,f_{N-1}(\mathbf{x}_i)]\}_{i=1}^M. 
\end{equation}

It is worth noting that more cross-stage representation integration methods can be further tailored to accommodate different downstream tasks in the future. In this work, we use a simple concatenation as a preliminary baseline to demonstrate MCL's main concept. In addition, since our framework does not change the backbone model, it can be seamlessly integrated with any NCE-based contrastive learning method.

\section{Experiments}
\label{sec:experiments}
\subsection{Datasets}
\textbf{Trifeature.} Trifeature~\cite{hermann2020shapes} is an image dataset, where each image has three independent features: color, texture and shape each taking 10 values. For each combination of the three features there are 100 samples, in which the position and rotation of the object are random. The three downstream tasks are to classify the color ($C=10$ classes), texture ($C=10$ classes), and shape ($C=10$ classes).

\noindent\textbf{CIFAR-MNIST (C-M).} CIFAR-MNIST consists of channel-wise concatenation of the CIFAR-10~\cite{krizhevsky2009learning} image and the MNIST~\cite{lecun2010mnist} image, following Chen \etal~\cite{chen2021intriguing}. Each image has four channels: three from the CIFAR-10 and one from the MNIST. As the images are randomly sampled from the two datasets in concatenation, the CIFAR-10 class and the MNIST class can be considered as two independent features. The two downstream tasks are to predict the CIFAR-10 class ($C=10$ classes) and the MNIST class ($C=10$ classes).

\noindent\textbf{CelebA.} CelebFaces Attributes Dataset (CelebA)~\cite{liu2018large} is a large-scale face attributes dataset, where each image has 40 attribute annotations. We take three attributes: black hair, male, and smiling. We resampled the dataset to make these three attributes independent of each other, which can be considered as three independent features. After resampling the dataset has more than 40k images in total. The three downstream tasks are to predict whether the celebrity has black hair ($C=2$ classes), whether the celebrity is smiling ($C=2$ classes), and whether the celebrity is male ($C=2$ classes).

\subsection{Baselines}
\textbf{IFM.} Implicit Feature Modification (IFM)~\cite{robinson2021can} aims to mitigate feature suppression by adaptive modifying samples to remove whichever features are used to discriminate a particular positive pair from negatives in feature space.

\noindent\textbf{FD.} Feature Dropout (FD)~\cite{tamkin2022feature} mitigates feature suppression by adversarial perturbation on input space to break the features already used to discriminate a particular positive pair from negatives, forcing the model to learn new features. 

\noindent\textbf{TS.} Temperature Schedules (TS)~\cite{kukleva2023temperature} aims to improve the contrastive learning performance on long-tail data by dynamically scheduling the temperature parameter along the training process. We consider it as a baseline here since temperature has a strong impact on which features are learned. 

\subsection{Evaluation}
\textbf{Linear Evaluation Protocol.} The well-accepted linear evaluation protocol~\cite{chen2020simple} is used to evaluate the quality of learned representations in the unimodal contrastive learning setting. Specifically, after the contrastive training process, the trained encoder is fixed and the projection head is discarded. A linear softmax classifier is trained on top of the trained encoder for each label. If the classification accuracy on the label related to one feature is high, it means the encoder encodes sufficient information about that feature, and vice versa.

\noindent\textbf{MMVP.} The Multimodal Visual Patterns (MMVP) benchmark~\cite{tong2024eyes} is used for the multimodal contrastive learning setting. It challenges the CLIP~\cite{radford2021learning} model to accurately match images with the corresponding text statements (\eg, ``a rabbit facing left'' and ``a rabbit facing right'') using the image-text similarity computed on learned representations and evaluates the pairing accuracy. 
Sourced from ImageNet~\cite{russakovsky2015imagenet} and LAION-Aesthetics~\cite{schuhmann2022laion}, the MMVP dataset comprises image pairs that exhibit high similarity in the CLIP embedding space but possess distinctly different semantic features in nine attributes (\eg, orientation and direction, color and appearance). 
Lower performance on the MMVP benchmark indicates a more severe issue of feature suppression.

\subsection{Mitigating Feature Suppression in the Unimodal Setting}
\label{exp:unimodal}
First, we compare our proposed MCL with IFM, TS, FD, and vanilla SimCLR on Trifeature, CelebA, and CIFAR-MNIST datasets.

\noindent\textbf{Experiment Setting.} We use SimCLR as the backbone contrastive learning method for MCL and all the baselines. For CIFAR-MNIST, we adapt ResNet-18 to accommodate smaller input sizes, following the CIFAR-10 configuration described by He \etal~\cite{he2016deep}. We choose the best temperature value from $\{0.1,0.25,0.5\}$ for MCL, IFM, FD, and vanilla SimCLR, and set the temperature schedule range for TS to $[0.1,1]$ as its default setting in comparison. For MCL, we train for $3$ stages, and $200$ epochs in each stage. The number of clusters for K-means is set to $5$. For IFM and FD, we train for $200$ epochs. For TS, we train for $600$ epochs since it requires a long training time in adherence to the requirements of the original setting. We use the default settings for all the baselines and backbone models unless otherwise specified.

\begin{figure}[htbp]
  \centering
  \begin{minipage}{0.6\textwidth}
    \centering
    \includegraphics[width=\linewidth]{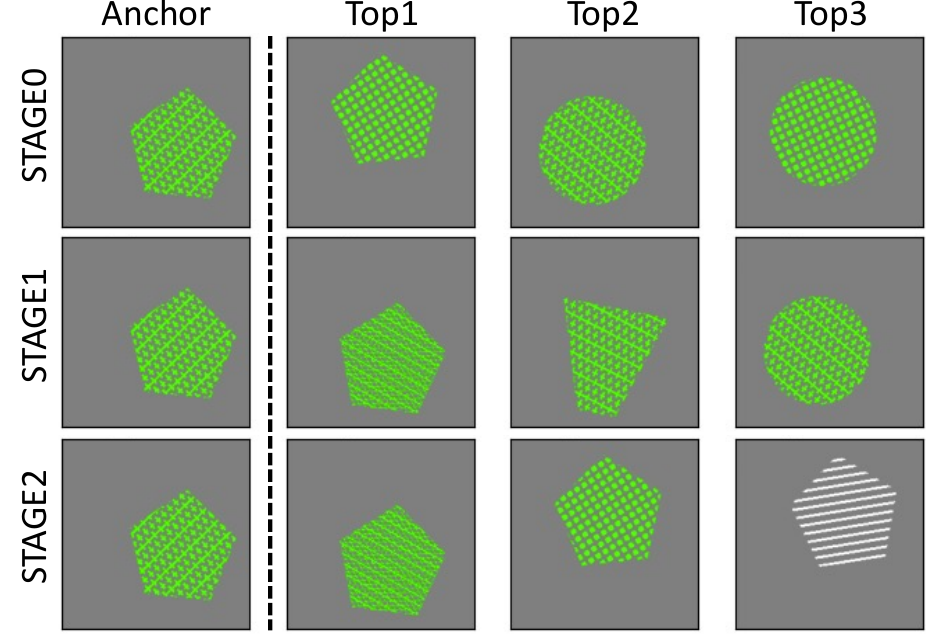}
    \caption{In each stage, the top 3 most similar samples to the anchor, which demonstrates the model’s shifting focus from color, texture, to shape across different stages of MCL.}
    \label{fig:visual_trifeature}
  \end{minipage}
  \hfill
  \begin{minipage}{0.36\textwidth}
    \centering
    \includegraphics[width=0.94\linewidth]{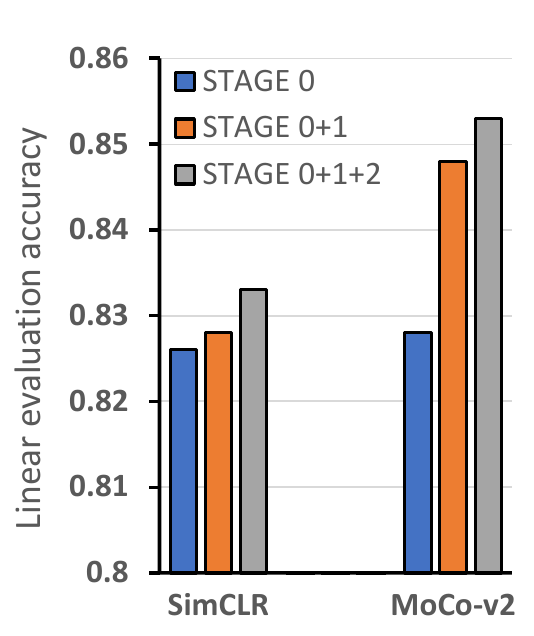}
    \caption{Linear evaluation accuracy on STL-10 dataset by incorporating MCL with SimCLR and MoCo-v2.}
    \label{fig:result_stl}
  \end{minipage}
\end{figure}

\noindent\textbf{Experimental Results.} As shown in \cref{tab:compare}, MCL achieves non-trivial improvements compared with the baselines in almost all the tasks. The performance gain is especially prominent in CIFAR-MNIST and Trifeature, where the feature suppression is severe. In CIFAR-MNIST, the CIFAR feature is entirely dominated by the MNIST feature. Whereas, with our MCL framework, the linear evaluation accuracy of the MNIST is largely improved. Meanwhile, the already well-learned features in vanilla SimCLR are maintained in MCL in all the settings, which indicates MCL can learn the previously ignored features without forgetting the already well-learned features. 
% 
% Compared with the other baselines, although they achieve some improvements on a few tasks, they suffer from performance degradation on the other tasks, leading to worse performance than the vanilla SimCLR on average. 
% 
Though the baselines may show improvements on certain tasks, they experience performance declines on others, leading to an overall performance that is inferior to the vanilla SimCLR.
% 
% MCL achieves non-trivial improvements compared with the three baseline methods. 
% 
% To analyze the dominant features in different stages, we demonstrate the Top-3 samples most similar to the anchor in the three stages, respectively. 
To illustrate MCL’s ability to prioritize different key features across stages, we present the three samples most similar to the anchor for each stage. The common characteristics of these top 3 samples highlight the model’s shifting focus throughout the learning process. For example, the similarity in shape between the anchor and the top 3 samples in Stage 2 indicates the model’s concentration on shape at this stage. From \cref{fig:visual_trifeature}, we can observe the model evolves through stages, with a prioritization on color in Stage 0, texture in Stage 1, and shape in Stage 2.

\begin{table}[ht]\footnotesize
    \centering
    \caption{Linear evaluation accuracy of MCL and the baselines. Bold indicates the best performance.}
    \begin{tabular}{l|c c c| c c c| c c| c}
    \toprule
     & \multicolumn{3}{c|}{Trifeature} & \multicolumn{3}{c|}{CelebA} & \multicolumn{2}{c|}{CIFAR-MNIST} &  \\
    \midrule 
     & Shape & Texture   & Color & Hair  & Smiling & Gender &   CIFAR   &   MNIST   &  Average\\
    \midrule 
     IFM    & 0.99  & 0.99   & \textbf{1.00} & 0.82  & 0.70 & 0.93 &   0.11   &   \textbf{0.99}   &  0.82\\
     TS    & 0.93 & \textbf{1.00}   & \textbf{1.00} & 0.62  & 0.65 & 0.87 &   0.09   &   \textbf{0.99}   &  0.77\\
     FD    & 0.75 & 0.73  & 0.78 & 0.81  & \textbf{0.89} & 0.94 &   0.78   &   0.86   &  0.82\\
     SimCLR& 0.81 & 0.99  & \textbf{1.00} & 0.84  & 0.75 & 0.94 &   0.29   &   0.98   &  0.83\\
     MCL    & \textbf{1.00} & \textbf{1.00}   & \textbf{1.00} & \textbf{0.85}  & 0.79 & \textbf{0.95} &   \textbf{0.87}   &   \textbf{0.99}   &  \textbf{0.93}\\
    \bottomrule  
    \end{tabular}
    \label{tab:compare}
\end{table}

% Different from the above datasets where the semantic features are already known, in more real-case datasets the features are hidden.

% Unlike the previously mentioned datasets where semantic features are explicitly identified, in more real-world datasets, these features are often unknown. Consequently, one will not be able to explicitly observe the potential feature suppression in real datasets.
Unlike  CIFAR-MNIST and Trifeature, STL-10~\cite{coates2011analysis} does not have explicitly identified semantic features.
% Whereas, incorporating MCL can still improve the model's performance in real datasets. 
We incorporate MCL with both SimCLR and MoCo-v2~\cite{chen2020improved}, use ResNet-50~\cite{he2016deep} as the encoder, train on STL-10 dataset for $3$ stages, $400$ epochs for each stage. The results are shown in \cref{fig:result_stl}. 
MCL boosts the MoCo-v2 performance on STL-10 by more than two percent.
% using two additional stages. 
This further validates the effectiveness of our proposed MCL framework.

\subsection{Mitigating Feature Suppression in the Multimodal Setting}
% \textbf{O\&D}: Orientation and Direction, \textbf{PSF}: Presence of Specific Features, \textbf{S\&C}: State and Condition, \textbf{Q\&C}: Quantity and Count, \textbf{P\&R}: Positional and Relational Context, \textbf{C\&A}: Color and Appearance, \textbf{S\&P}: Structural and Physical Characteristics, \textbf{Texts}: Texts, \textbf{V\&P}: Viewpoint and Perspective.
\begin{table}[ht]\footnotesize
    \centering
    \caption{Accuracy of the MCL tuned CLIP models on the MMVP benchmark. Bold indicates the best performance across four stages. The STAGE0 model is the original OpenAI pre-trained model. The final MCL model result is obtained by using the concatenation of the representations across all four stages as the final representation.}
    \begin{tabular}{l|c c c c c c c c c c}
    \toprule
     & \textbf{O\&D} & \textbf{PSF} & \textbf{S\&C} & \textbf{Q\&C} & \textbf{P\&R} & \textbf{C\&A} & \textbf{S\&P}   & \textbf{Texts}  & \textbf{V\&P} & Average\\
    \midrule 
     ResNet STAGE0& 6.7 & 6.7  & \textbf{46.7} & 0.0  & \textbf{13.3} & 46.7 &   \textbf{33.3}   &   6.7   &  13.3 & 19.3\\
     ResNet STAGE1& \textbf{13.3} & \textbf{13.3}  & 40.0 & \textbf{20.0}  & 0.0 & 33.3 &   33.3   &   6.7   &  26.7 & 20.7\\
     ResNet STAGE2& 6.7 & 6.7  & 33.3 & 6.7  & 6.7 & \textbf{73.3} &   13.3   &   13.3   &  \textbf{33.3} & 21.5\\
     ResNet STAGE3& 0.0 & 13.3  & 33.3 & 20.0  & 0.0 & 60.0 &   33.3   &   \textbf{26.7}   &  20.0 & 23.0\\
     ResNet MCL& 0.0 & 13.3  & 40.0 & 6.67  & 6.67 & 73.3 &   40.0   &   6.7   &  33.3 & \underline{24.4}\\
     \midrule 
     ViT STAGE0& \textbf{26.7} & 13.3  & 26.7 & 6.7  & 6.7 & 40.0 &   26.7   &   \textbf{13.3}   &  20.0 & 20.0\\
     ViT STAGE1& 13.3 & \textbf{33.3}  & \textbf{66.7} & 26.7  & \textbf{13.3} & 53.3 &   20.0   &  13.3   &  \textbf{26.7} &29.6\\
     ViT STAGE2& 0.0 & 13.3  & 46.7 & \textbf{40.0}  & 6.7 & 53.3 &   20.0   &  13.3   &  20.0 &23.7\\
     ViT STAGE3& 13.3 & 6.67  & 46.7 & 13.3  & 13.3 & \textbf{66.7} &   \textbf{40.0}   &  13.3   &  20.0 &25.9\\
     ViT MCL& 6.67 & 20.0  & 73.3 & 13.3  & 13.3 & 80.0 &  46.7  &  13.3   &  26.7 & \underline{32.6}\\
    \bottomrule  
    \end{tabular}
    \label{tab:clip}
\end{table}

% \noindent\textbf{CC12M.} Conceptual 12M (CC12M)~\cite{changpinyo2021conceptual} is a dataset with 12 million image-text pairs specifically meant to be used for vision-and-language pre-training.

\noindent\textbf{Experiment Setting.}
In this setting, we mainly adopt CLIP~\cite{radford2021learning} to learn image representations on image-text pair datasets using contrastive learning. Specifically, we train two CLIP models in the MCL framework with different scales: one uses ResNet-50 as the image encoder, while the other utilizes ViT-L-14~\cite{radford2021learning}. Since the CLIP model requires training on hundreds of GPUs for a few days, instead of training from scratch, we tune the original CLIP model on Conceptual 12M (CC12M)~\cite{changpinyo2021conceptual}, a large image-text pair dataset specifically designed for vision-and-language pre-training. We use the original OpenAI pre-trained weights\etal~\cite{radford2021learning} as the initialization for each stage. 
% and default settings for all the baselines and backbone models unless otherwise specified. 
Our implementation is based on OpenCLIP~\cite{cherti2023reproducible}. 
% For the ViT-L-14, since the model is large and CC12M is relatively tiny compared to the training dataset of CLIP, we fixed part of the model weights to avoid overfitting. 
Since we mainly focus on image representation, we fix the text encoder in contrast to the approach in Zhai \etal~\cite{zhai2022lit}. Considering the features of the last few blocks in CLIP are more dominant~\cite{gandelsman2023interpreting} and avoid overfitting, we only leave the last $6$ blocks trainable, with the other part of the image encoder fixed. 
For both two versions, we tune the models on CC12M for additional $3$ stages, $10$ epochs in each stage for the ResNet version, and $20$ epochs for the ViT version. We do K-means clustering on the image representation between each stage, and the number of clusters is set to $10$. We use a batch size of $8192$ and a warmup of $10\%$ of the total steps. It is worth noting that although multiple stages are trained in our framework, the computation parameters will not increase much. For example, the total number of parameters in our four-stage ViT CLIP image encoder will only increase by 75\% compared to a single vanilla CLIP. The parameters can be further reduced by training fewer layers or introducing parameter-efficient tuning, such as LoRA~\cite{hu2021lora}. We leave this as future work. 

% We also demonstrate the performance of the ensemble model of the four stages after cross-stage representation integration, which can be considered as a performance lower bound. 
% By contrast, in the linear evaluation protocol, the linear classification layer can implicitly assign different weights for representations from different stages.

%need revision
\noindent\textbf{Experimental Results.}
\Cref{tab:clip} presents the performance of the tuned CLIP models across different stages. For both the ResNet and ViT architectures, we observe improvements in later stages over earlier ones on tasks where initial performances were suboptimal. For example, ViT at Stage 1 outperforms its Stage 0 counterpart in \textbf{PSF}, ViT at Stage 2 exceeds the performance of both Stage 0 and Stage 1 in \textbf{Q\&C}, and ViT at Stage 3 surpasses the earlier stages in \textbf{C\&A}. These improvements underscore the models’ evolving expertise in distinct features, enhancing their competence with specific attributes.
Notice that each stage of the ViT model specializes in different attributes: ViT-STAGE0 in \textbf{O\&D}, ViT-STAGE1 in \textbf{PSF}, \textbf{S\&C}, and \textbf{V\&P}, ViT-STAGE2 in \textbf{Q\&C}, and ViT-STAGE3 in \textbf{C\&A} and \textbf{S\&P}. 

Upon integrating representations from all stages, the ensemble model’s average performance on the MMVP benchmark increases from 19.3 to 24.4 for the ResNet version, and from 20.0 to 32.6 for the ViT version. Notably, the final MCL model surpasses the peak performance seen in individual stages in certain attributes (\eg, \textbf{S\&C} and \textbf{C\&A} for ViT), suggesting that it captures complementary features across stages.
Interestingly, in certain attributes (\eg, \textbf{O\&D}), the MCL model does not achieve the best results compared to individual stages, suggesting room for improvement in our approach of cross-stage representation integration. This finding directs future research toward optimizing the process of combining stage-specific insights, aiming to harness the full potential of the model’s learned features.

\subsection{Discussion}
In this section, 
% we delve into an analysis of MCL’s learning dynamics. Our focus includes understanding the progression of learning distinct features across 
% examining the impact of varying the 
we study the impact of training stage ($N$), number of clusters ($K$) in K-means, and temperature $\tau$ in our MCL framework. Unless noted otherwise, the experimental configurations adhere to those outlined in \cref{exp:unimodal}.

\begin{figure}[htbp]
  \centering
  \begin{minipage}{0.58\textwidth}
    \begin{subfigure}[l]{0.48\textwidth}
        \includegraphics[width=\textwidth]{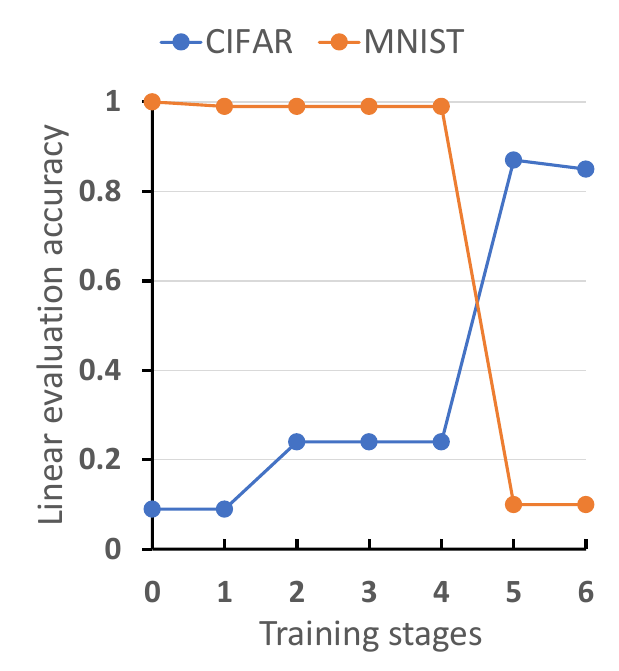}
        \caption{Linear evaluation accuracy for each stage individually before cross-stage representation integration.}
        \label{fig:stage}
    \end{subfigure}
    \hfill
    \begin{subfigure}[l]{0.48\textwidth}
        \includegraphics[width=\textwidth]{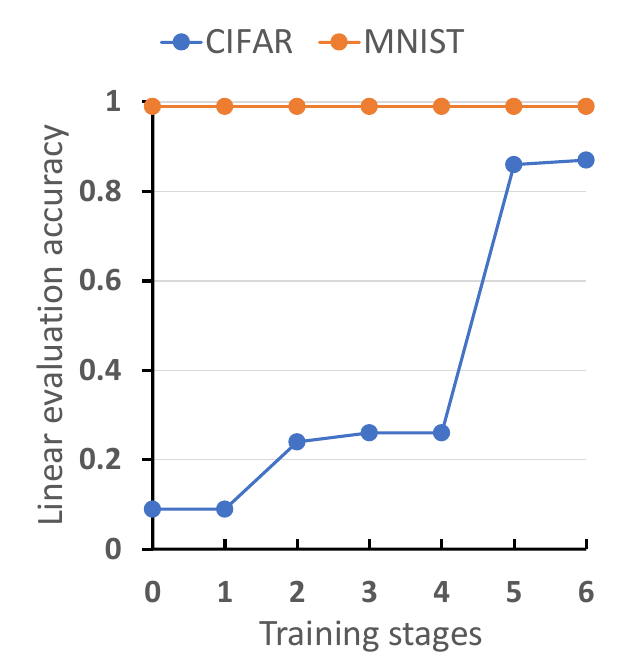}
        \caption{Linear evaluation accuracy for training $N$ stages after cross-stage representation integration.}
        \label{fig:stage_total}
    \end{subfigure}
    \caption{Linear evaluation results of MCL on CIFAR-MNIST for different training stages.}
  \end{minipage}
  \hfill % 添加一些水平空间
  \begin{minipage}{0.38\textwidth}
    \centering
    \includegraphics[width=0.8\linewidth]{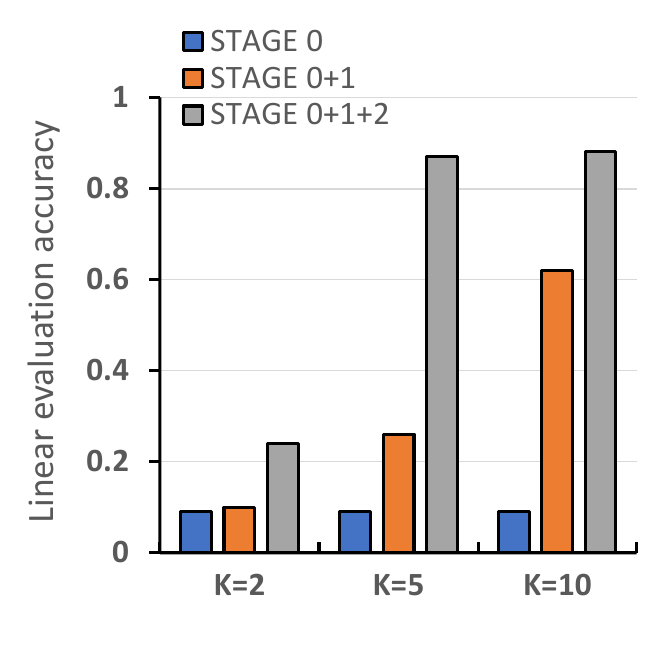}
    \caption{Linear evaluation results of MCL on CIFAR using $K=\{2,5,10\}$, trained on CIFAR-MNIST for three stages respectively.}
    \label{fig:K}
  \end{minipage}
\end{figure}

\noindent\textbf{Learning Dynamics Across Stages.}
We explore the learning dynamics by increasing the number of training stages from one to seven. Given the constraint in \cref{eq:kn_bound}, a larger number of stages ($N$) requires a smaller number of clusters ($K$). Therefore, we set $K=2$ and the temperature $\tau=0.25$. Initially, we assess the performance of the encoder at each stage independently, without cross-stage representation integration. As depicted in \cref{fig:stage}, up to Stage 4, CIFAR features remain predominantly suppressed by MNIST features. Interestingly, at Stage 5, the model abruptly shifts to prioritize CIFAR features over MNIST, resulting in a sudden change rather than a gradual improvement in linear evaluation accuracy. Subsequently, we analyze the performance with cross-stage representation integration across $N$ stages, illustrated in \cref{fig:stage_total}. This process effectively preserves the initially learned MNIST features. Without cross-stage integration, there is a noticeable drop in MNIST feature performance from Stage 4 to 5. However, with integration, the performance on MNIST remains stable. From Stage 5 onwards, the model exhibits minimal performance variation indicating convergence.

\noindent\textbf{Impact of Number of Clusters.}
We investigate how the number of clusters $K$ in k-means clustering affects the learning process within the MCL framework. We train SimCLR in the MCL framework on CIFAR-MNIST for three stages, varying $K$ among $\{2,5,10\}$ respectively and assess the linear evaluation accuracy on CIFAR after cross-stage representation integration of $N=\{0,1,2\}$ stages, given CIFAR features were notably suppressed in the vanilla SimCLR (STAGE0). The findings, depicted in \cref{fig:K}, indicate that a higher $K$ value enables the model to uncover previously suppressed features more rapidly, reducing the number of stages needed to achieve substantial performance gains. Conversely, a lower $K$ value ($K=2$) necessitates additional stages for similar outcomes, as evidenced in \cref{fig:stage_total}. Nonetheless, increasing cluster count from 5 to 10 does not markedly improve performance beyond the third stage, suggesting a performance plateau. Further increasing $K$ beyond this point does not yield additional benefits.

\noindent\textbf{Robustness to Diverse Temperature Settings.}
Following~\cite{robinson2021can}, which highlight the pivotal role of temperature ($\tau$) in influencing feature suppression within contrastive learning frameworks, we evaluate MCL’s adaptability across varying temperature settings $\tau=\{0.1,0.25,0.5\}$. Results, as detailed in \cref{tab:multitau}, indicate that MCL consistently enhances performance across all datasets, irrespective of the temperature setting employed. Notably, the most significant performance uplifts are observed under temperature conditions where the baseline SimCLR model exhibits pronounced feature suppression, such as $\tau=\{0.1,0.5\}$ in Trifeature. This trend underscores MCL’s capability to not only mitigate feature suppression but also fortify the robustness of SimCLR against temperature variations.
\begin{table*}[!ht]\footnotesize
    \centering
    \caption{The performance improvements by incorporating MCL with SimCLR under different temperature settings. Improvements are observed across all datasets, with more significant gains where feature suppression is more pronounced. }
    \begin{tabular}{l|c c| c c| c c}
    \toprule
     & \multicolumn{2}{c|}{$\tau=0.1$} & \multicolumn{2}{c|}{$\tau=0.25$} & \multicolumn{2}{c}{$\tau=0.5$}\\
    \midrule 
     & SimCLR & MCL   & SimCLR & MCL   & SimCLR & MCL\\
    \midrule 
     Trifeature(Shape)   & 0.66 & 1.00 ($\uparrow$.34) & 0.81 & 0.92 ($\uparrow$.11)   & 0.44 & 0.75 ($\uparrow$.31) \\
     Trifeature(Texture) & 0.91 & 1.00 ($\uparrow$.09) & 0.99 & 1.00 ($\uparrow$.01)   & 0.92 & 0.99 ($\uparrow$.07) \\
     Trifeature(Color)   & 1.00 & 1.00 ($\uparrow$.00) & 1.00 & 1.00 ($\uparrow$.00)   & 1.00 & 1.00 ($\uparrow$.00) \\
     \midrule 
     CelebA(Hair)    & 0.84 & 0.85 ($\uparrow$.01)   & 0.58 & 0.73 ($\uparrow$.15)   & 0.57 & 0.69 ($\uparrow$.12) \\
     CelebA(Smiling) & 0.75 & 0.79 ($\uparrow$.04)   & 0.63 & 0.65 ($\uparrow$.02)   & 0.63 & 0.66 ($\uparrow$.03) \\
     CelebA(Gender)  & 0.94 & 0.95 ($\uparrow$.01)   & 0.72 & 0.87 ($\uparrow$.15)   & 0.71 & 0.86 ($\uparrow$.15) \\
     \midrule 
     C-M(CIFAR) & 0.29 & 0.83 ($\uparrow$.54)   & 0.10 & 0.87 ($\uparrow$.77)   & 0.10 & 0.87 ($\uparrow$.77) \\
     C-M(MNIST) & 0.98 & 0.98 ($\uparrow$.00)   & 0.99 & 0.99 ($\uparrow$.00)   & 0.99 & 0.99 ($\uparrow$.00)\\
    \bottomrule  
    \end{tabular}
    \label{tab:multitau}
\end{table*}

\section{Conclusion}
% In this study, we mitigated the issue of feature suppression present in both unimodal and multimodal settings of contrastive learning. We introduced the Multistage Contrastive Learning (MCL) framework, a novel approach designed to alleviate feature suppression. Our framework’s cross-stage negative sampling strategy successfully encourages the learning of previously unlearned information at each stage. We demonstrated that by progressing through stages and integrating representations across stages, MCL effectively preserves well-learned features and mitigates degradation observed in prior works. The superiority of our approach is clearly illustrated through comparative analyses with baseline models across a variety of datasets and settings, highlighting MCL’s robustness in both unimodal and multimodal contrastive learning contexts.

In this paper, we investigated the critical feature suppression in contrastive learning. Specifically, we introduced the Multistage Contrastive Learning (MCL) framework, a novel, model-agnostic framework. MCL employs a cross-stage negative sampling strategy that effectively promotes the learning of previously unlearned information at each stage.
Meanwhile, 
% by progressing through stages and integrating representations across stages, 
MCL efficiently preserves well-learned features and mitigates degradation observed in prior works. The effectiveness of our approach is demonstrated through comprehensive analyses with commonly used baseline models on various datasets and settings, highlighting MCL's effectiveness and adaptability in both unimodal and multimodal contrastive learning.
\clearpage

\section*{Acknowledgement}
% \noindent\textbf{Acknowledgement.}
This project was funded by the National Research Foundation Singapore under AI Singapore Programme (Award Number: AISG-GC-2019-001-2B and AISG2-TC-2022-004).

\bibliographystyle{splncs04}
\bibliography{main}

\clearpage
\appendix
\section{Appendix}
In the appendix, we provide further details and additional experimental results that complement the main text. The contents include:

\Cref{sec:1} . Implementation details of \textit{feature-aware negative sampling}.

%\item Additional experiment of fine-tuning ResNet CLIP model pretrained on CC12M using MCL (\Cref{sec:2}).

\Cref{sec:3} . Learned features hinder the model from acquiring new features.

\Cref{sec:4} . Train SimCLR using MCL on ImageNet.

\Cref{sec:5} . Ensemble CLIP models trained using MCL with Model Soup.

\Cref{sec:6} . The computational cost of MCL.

\Cref{sec:7} . Ablation study for the number of trainable blocks when training ViT CLIP models using MCL.

\Cref{sec:8} . Analysis of the clustering results across different stages of MCL.

\subsection{Details of Feature-Aware Negative Sampling}
\label{sec:1}
The \textit{feature-aware negative sampling} strategy necessitates that an anchor and its negative samples belong to the same cluster, as determined in the previous stages. To effectively organize training samples according to their pseudo labels obtained between stages, we utilize a custom batch sampler. The implementation details of the batch sampler are shown in \cref{alg}.

\begin{algorithm}
\caption{Feature-Aware Negative Sampling for MCL}
\label{alg}
\begin{algorithmic}[1]
\State \textbf{Input:} Dataset $D$ with pairs of data and pseudo labels $\{(x_i,y_i)\}_{i=1}^M$, number of unique pseudo labels $C = |\{y_i\}_{i=1}^M|$
\State Group $D$ into a two-dimensional array $\texttt{G}$ based on pseudo labels $y$, such that $\texttt{G}[c] = \{(x_i,y_i) \in D | y_i = c\}$
\For{batch index $j \gets 0$ \textbf{to} $\infty$} 
    \State Select group index $k \gets j \mod C$
    \If{all samples in $\texttt{G}[k]$ are traversed}
        \State \textbf{continue}
    \EndIf
    \State Create batch from untraversed samples in $\texttt{G}[k]$
    \If{all samples in $\texttt{G}$ are traversed}
        \State \textbf{break}
    \EndIf
\EndFor
\end{algorithmic}
\end{algorithm}

\subsection{Learned Features Hinder the Learning Process}
\label{sec:3}
% To investigate this, we conduct an inheritance experiment by initializing the model from the last stage instead of training from scratch.
We conduct experiments to demonstrate that the learned features can impede the model's ability to acquire new features. Specifically, at each stage, we initialize the model with parameters from the previous stage instead of training from scratch.
We train SimCLR with ResNet18 as the encoder on the Trifeature dataset. We use the $\tau=0.5$ setting as described in Section 5.6 in the main paper.
As shown in \cref{tab:inherit}, this model makes almost no progress from Stage 1 to Stage 2, and its final performance (0.42/0.97/1.00) is markedly inferior to that of our proposed method (0.75/0.99/1.00). 
% and its final performance (0.42/0.97/1.00) is significantly worse compared to our proposed method (0.75/0.99/1.00). 
% 
This can be attributed to the design of MCL, which does not require the retention of previously learned properties at each stage, thus allowing it to freely learn new features without constraints.
% In MCL, the model at each stage is not required to retain previously learned properties, allowing it to freely learn new features without constraints.

\begin{table}[ht]
    \centering
    \setlength{\tabcolsep}{5pt}
    \footnotesize
    \caption{Linear evaluation results on shape, texture, and color respectively using naive inheritance and MCL. }
    \begin{tabular}{l|c c c  }
    \toprule
      & Stage0  & Stage1 &Stage2   \\
    \midrule 
    Inheritance     & 0.44/0.92/1.00    &0.40/0.97/1.00      &0.42/0.97/1.00 \\
    MCL & 0.44/0.92/1.00    &0.72/0.97/1.00      &0.75/0.99/1.00 \\
    \bottomrule  
    \end{tabular}
    \label{tab:inherit}
\end{table}

\subsection{Train SimCLR Using MCL on ImageNet}
\label{sec:4}
 To further validate the effectiveness of MCL on the large unimodal dataset, we train SimCLR with ResNet34 as the encoder on ImageNet. We use a batch size of 8192 and train for 100 epochs at each stage for 3 stages. As shown in \cref{tab:imagenet}, the experiment results demonstrate a notable improvement. 
\begin{table}[ht]
    \centering
    \footnotesize
    \setlength{\tabcolsep}{5pt}
    \caption{Linear evaluation accuracy of incorporating MCL with SimCLR on ImageNet.}
    \begin{tabular}{l|c c c  }
    \toprule
      &Stage0 & Stage0+1   & Stage0+1+2 \\
    \midrule 
    MCL    &0.456   &0.477   & 0.482  \\
    \bottomrule  
    \end{tabular}
    \label{tab:imagenet}
\end{table}

\subsection{Ensemble ViT CLIP Models with Model Soup}
\label{sec:5}
% Model Soup~\cite{wortsman2022model} is the ensemble model obtained by averaging the weights of multiple fine-tuned models. This approach is straightforward yet efficient for model ensembling. 
% Model Soup~\cite{wortsman2022model} is an ensemble model created by averaging the weights of multiple fine-tuned models. This method is both simple and effective for model ensembling.
% 
% We applied this approach to ensemble the fine-tuned ViT CLIP models across four stages described in Section 5.4 of our main text. This method allows us to compare MCL fine-tuned models with the vanilla OpenAI pretrained model under the same parameter size. 
We conduct experiments to compare MCL fine-tuned models with the standard OpenAI pretrained model, maintaining identical parameter sizes. Specifically, we utilize Model Soup~\cite{wortsman2022model}, an effective and efficient ensemble approach that averages the weights of multiple fine-tuned models. We applied Model Soup to ensemble the fine-tuned ViT CLIP models across four stages, as described in Section 5.4 of the main text.
% 
% As shown in \cref{tab:fusion}, the MCL Model Soup attains an average score of 29.6, which, despite a minor performance reduction compared to the simple concatenation method discussed in our main text, still significantly outperforms the baseline average score of 20.0 with identical parameter size. 
As illustrated in \cref{tab:fusion}, the MCL Model Soup achieves an average score of 29.6. Although this represents a slight decrease in performance compared to the simple concatenation method discussed in our main text, it still substantially surpasses the baseline average score of 20.0, with the same parameter size.
We also test the MCL Model Soup on text-to-image and image-to-text retrieval tasks, following the settings in Zhang \etal~\cite{zhang2024long}. As shown in \cref{tab:fusion2}, the MCL Model Soup achieves a notable performance improvement compared to the OpenAI baseline. The results underscore the robustness and effectiveness of the MCL framework.

\begin{table}[ht]\footnotesize
    \centering
    \caption{Comparative analysis of ViT CLIP models fine-tuned with MCL on the MMVP benchmark. ``OpenAI'' denotes the baseline performance using the OpenAI pretrained model. ``MCL Concat'' represents the performance after applying MCL and using concatenated representations from all stages. MCL Model Soup (denoted as ``MCL MS'') illustrates the performance of the ensemble model created by Model Soup.}
    \begin{tabular}{l|c c c c c c c c c c}
    \toprule
     & \textbf{O\&D} & \textbf{PSF} & \textbf{S\&C} & \textbf{Q\&C} & \textbf{P\&R} & \textbf{C\&A} & \textbf{S\&P}   & \textbf{Texts}  & \textbf{V\&P} & Average\\
    \midrule 
     OpenAI& 26.7 & 13.3  & 26.7 & 6.7  & 6.7 & 40.0 & 26.7  & 13.3  &  20.0 & 20.0\\
     MCL Concat & 6.7 & 20.0  & 73.3 & 13.3  & 13.3 & 80.0 &   46.7   &  13.3 &  26.7 & 32.6\\
     MCL MS & 13.3 & 26.7  & 46.7 & 20.0 & 6.7 & 60.0 &   26.7   &  33.3   &  33.3 & 29.6\\
    \bottomrule  
    \end{tabular}
    \label{tab:fusion}
\end{table}

\begin{table}[ht]
    \centering
    \caption{Performance of MCL Model Soup. Tasks include text-to-image (T2I) retrieval and image-to-text (I2T) retrieval on 5k COCO validation set and 30k Flickr30k dataset. We use top-1, top-5, and top-10 Recall (R@1, R@5, R@10) as the evaluation metrics.}
    \begin{tabular}{l|c c c |c c c |c c c |c c c }
    \toprule
    & \multicolumn{6}{c|}{COCO} & \multicolumn{6}{c}{Flickr30k}\\
    & \multicolumn{3}{c|}{Image-to-Text} & \multicolumn{3}{c|}{Text-to-Image}& \multicolumn{3}{c|}{Image-to-Text} & \multicolumn{3}{c}{Text-to-Image}\\
      & R@1 &R@5 &R@10 & R@1 &R@5 &R@10 & R@1 &R@5 &R@10 & R@1 &R@5 &R@10   \\
    \midrule 
    OpenAI     & 56.1&79.5&86.8    &35.4&60.1&70.2      &48.5&72.6&80.8  &28.0&49.3&58.7 \\
    MCL MS &60.4&82.4&89.5 &42.2&66.9&76.2 &52.7&76.6&84.4&36.5&59.0&68.1\\
    \bottomrule  
    \end{tabular}
    \label{tab:fusion2}
\end{table}

\subsection{The Computational Cost of MCL}
\label{sec:6}
% The computational cost of MCL is shown in \cref{tab:complexity}, using our largest model in unimodal (as in \cref{tab:imagenet}) and multimodal settings as examples. Notice that the fine-tuning cost of CLIP ViT is only about 3\% of the OpenAI pre-training cost in TFLOPs, and the inference cost is only increased by 75\%.
The computational cost of MCL is detailed in \cref{tab:complexity}, using our largest model in unimodal (as described in \cref{sec:4}) and multimodal settings for illustration. Notably, the fine-tuning cost of the CLIP model is approximately 3\% of the OpenAI pre-training cost in TFLOPs, and the inference cost increases by only 75\%.

\begin{table}[ht]
    \centering
    \caption{Computational complexity of MCL models. The training cost is evaluated by TFLOPs, Wall Clock Time (WCT), and TFLOPS. Inference cost is evaluated by relative complexity ratio (R) compared to backbone models (\ie SimCLR ResNet34 and CLIP ViT). WCT is measured on 8 A100 GPUs.}
    \begin{tabular}{l|c c c c }
    \toprule
      % &\multicolumn{3}{c|}{Training} & Inference   &     &\multicolumn{3}{c}{Training} & Inference \\ \midrule 
    & TFLOPs   & WCT    & TFLOPS & R \\
    \midrule 
    MCL SimCLR ResNet34 & $1.3\times 10^{7}$ & 30h & $1.6\times 10^{2}$ & 300\% \\
    MCL CLIP ViT & $1.8\times 10^{8}$ & 32h & $1.6\times 10^{3}$ & 175\% \\
    \bottomrule  
    \end{tabular}
    \label{tab:complexity}
\end{table}

\subsection{Exploring the Number of Trainable Blocks}
\label{sec:7}
% To understand the impact of different numbers of trainable blocks in the ViT architecture, we experimented with leaving the last 4, 6 (as detailed in the main text), and 8 transformer blocks as trainable. Due to computational constraints, we limited our exploration to the first three stages of MCL. All the other settings are identical to our main text. 
To assess the impact of varying the number of trainable blocks in the ViT architecture, we conduct experiments with the last 4, 6 (as detailed in the main text), and 8 transformer blocks set as trainable. Due to computational constraints, we focus on the first three stages of MCL. All other settings remained consistent with those described in the main text.
% 
% As summarized in \cref{tab:blocks}, making 4 blocks trainable achieves an average score of 28.1, outperforming the 6-block configuration (27.4 average score). However, extending to 8 trainable blocks reduces the average score to 25.9. Despite these variations, MCL consistently enhances performance over the vanilla OpenAI pretrained ViT CLIP, which scores an average of 20.0.
As summarized in \cref{tab:blocks}, configuring 4 blocks as trainable results in an average score of 28.1, surpassing the 6-block configuration, which achieves a 27.4 average score. However, extending to 8 trainable blocks decreases the average score to 25.9. Despite these variations, MCL consistently improves performance compared to the vanilla OpenAI pretrained ViT CLIP, which scores an average of 20.0.

\begin{table}[ht]\footnotesize
    \centering
    \caption{Performance comparison of MCL-tuned ViT CLIP models with different numbers of trainable blocks on the MMVP benchmark. ``S0'' denotes the original OpenAI pre-trained ViT CLIP model. ``S1'' and ``S2'' denote Stage 1 and Stage 2. ``MCL'' represents the results derived from concatenating representations across the three stages.}
    \begin{tabular}{c|l|c c c c c c c c c c}
    \toprule
    \makecell[c]{Trainable\\Blocks} & & \textbf{O\&D} & \textbf{PSF} & \textbf{S\&C} & \textbf{Q\&C} & \textbf{P\&R} & \textbf{C\&A} & \textbf{S\&P}   & \textbf{Texts}  & \textbf{V\&P} & Average\\
    \midrule 
     -&S0& 26.7 & 13.3  & 26.7 & 6.7  & 6.7 & 40.0 &   26.7   &   13.3   &  20.0 & 20.0\\
     \midrule
     \multirow{3}{*}{4} &S1& 0.0 & 20.0  &40.0 & 20.0  & 6.7 & 73.3 &   26.7   &  6.7   &  13.3 &23.0\\
     &S2& 20.0 & 20.0  & 40.0 & 13.3  & 20.0 & 73.3 &   26.7   &  0.0   &  20.0 &25.9\\
     &MCL& 13.3 & 33.3  & 60.0 & 20.0  & 13.3 & 73.3 &  13.3  &  6.7   &  20.0 &28.1\\
     \midrule
     \multirow{3}{*}{6} &S1& 13.3 & 33.3  & 66.7 & 26.7  & 13.3 & 53.3 &   20.0   &  13.3   &  26.7 &29.6\\
     &S2& 0.0 & 13.3  & 46.7 & 40.0  & 6.7 & 53.3 &   20.0   &  13.3   &  20.0 &23.7\\
     &MCL& 6.7 & 13.3  & 53.3 & 26.7  & 13.3 & 66.7 &  40.0  &  0.0   &  26.7 & 27.4\\
     \midrule
     \multirow{3}{*}{8} &S1& 13.3 & 20.0  & 46.7 & 6.7  & 13.3 & 66.7 &   20.0   &  20.0   &  20.0 &25.2\\
     &S2& 6.7 & 33.3  & 40.0 & 0.0  & 13.3 & 53.3 &   6.7   &  13.3   &  20.0 &20.7\\
     &MCL& 13.3 & 33.3  & 33.3 & 13.3  & 6.7 & 60.0 &  26.7  &  20.0   &  26.7 & 25.9\\
    \bottomrule  
    \end{tabular}
    \label{tab:blocks}
\end{table}

\subsection{Clustering Analysis}
\label{sec:8}

\begin{table}[ht]\footnotesize
    \centering
    \caption{AMI scores between K-means clustering outcomes across different MCL stages. Low AMI scores between distinct stages suggest substantially different clustering results.}
    \begin{tabular}{c|ccc}
    \toprule
        AMI & Stage 0 & Stage 1 & Stage 2\\
        \midrule 
         Stage 0& 1.00 & 0.33 & 0.25\\
         Stage 1& 0.33 & 1.00 & 0.24\\
         Stage 2& 0.25 & 0.24 & 1.00\\
         \bottomrule  
    \end{tabular}
    \label{tab:AMI}
\end{table}

% In this section, we delve into the clustering dynamics observed while training ViT CLIP models via the MCL approach, as detailed in Section 5.4 of the main text.
In this section, we explore the clustering dynamics observed during the training of ViT CLIP models using the MCL approach, as detailed in Section 5.4 of the main text.
We begin by examining the Adjusted Mutual Information (AMI)~\cite{vinh2009information} scores computed between the clustering outcomes of subsequent stages in the experiment. AMI scores range from 0, indicating no mutual information (independent clusterings), to 1, denoting identical clustering results.

As illustrated in \cref{tab:AMI}, the low AMI scores between distinct stages highlight the divergent clustering outcomes, underscoring that each stage learns a unique feature distribution. This divergence indicates the MCL framework's effectiveness in guiding the model to capture distinct, non-redundant features across stages.

\begin{figure}[htbp]
  \centering
    \begin{subfigure}[l]{0.32\textwidth}
        \includegraphics[width=\textwidth]{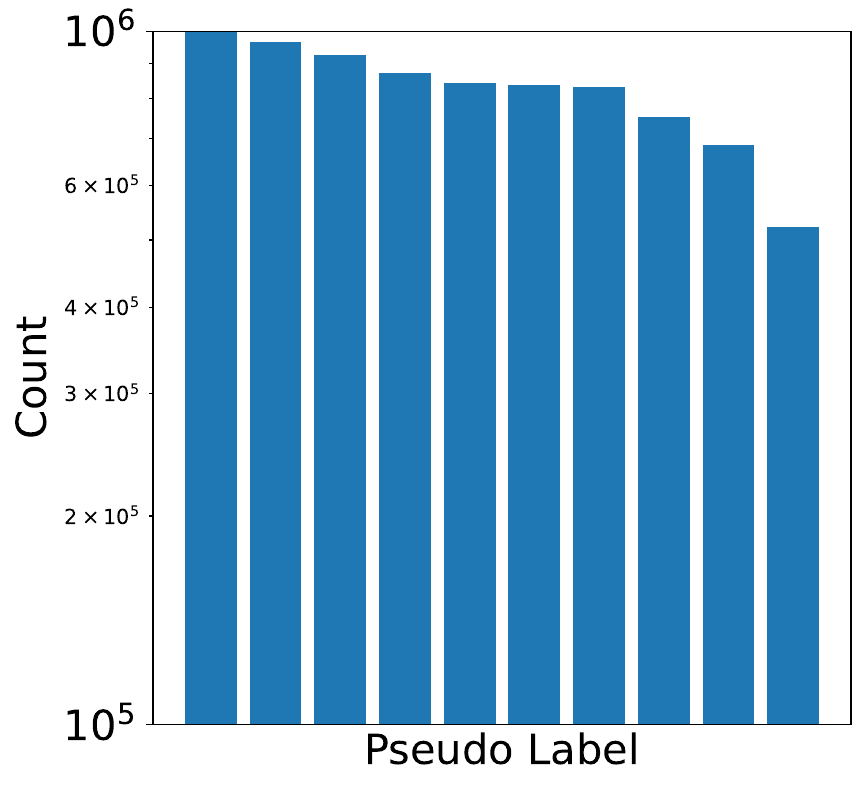}
        \caption{Stage 0}
    \end{subfigure}
    \hfill
    \begin{subfigure}[l]{0.32\textwidth}
        \includegraphics[width=\textwidth]{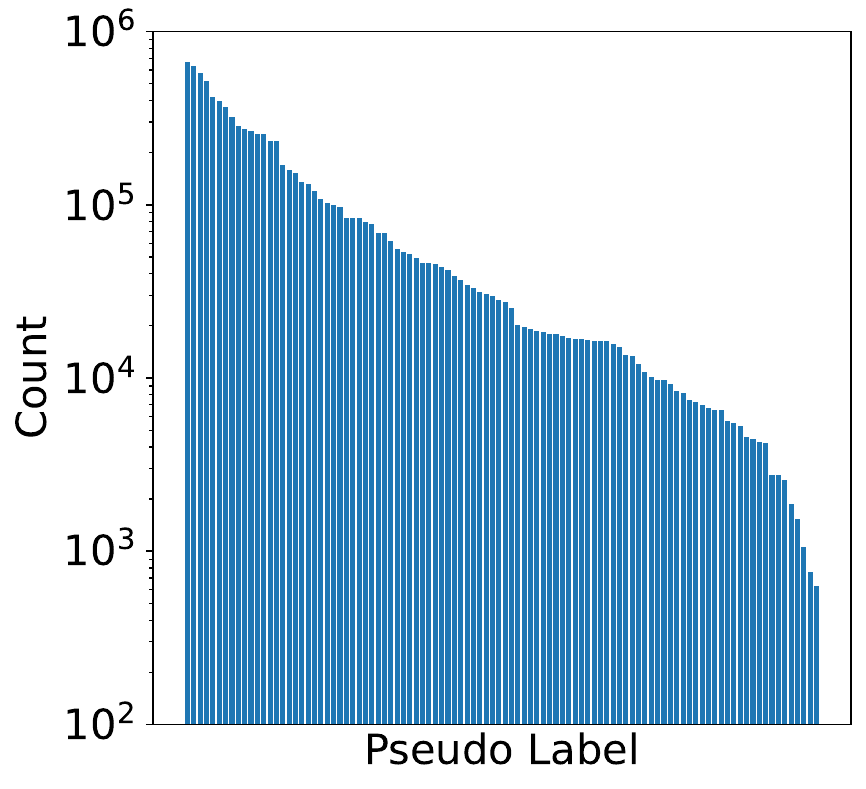}
        \caption{Stage 1}
    \end{subfigure}
    \hfill
    \begin{subfigure}[l]{0.32\textwidth}
        \includegraphics[width=\textwidth]{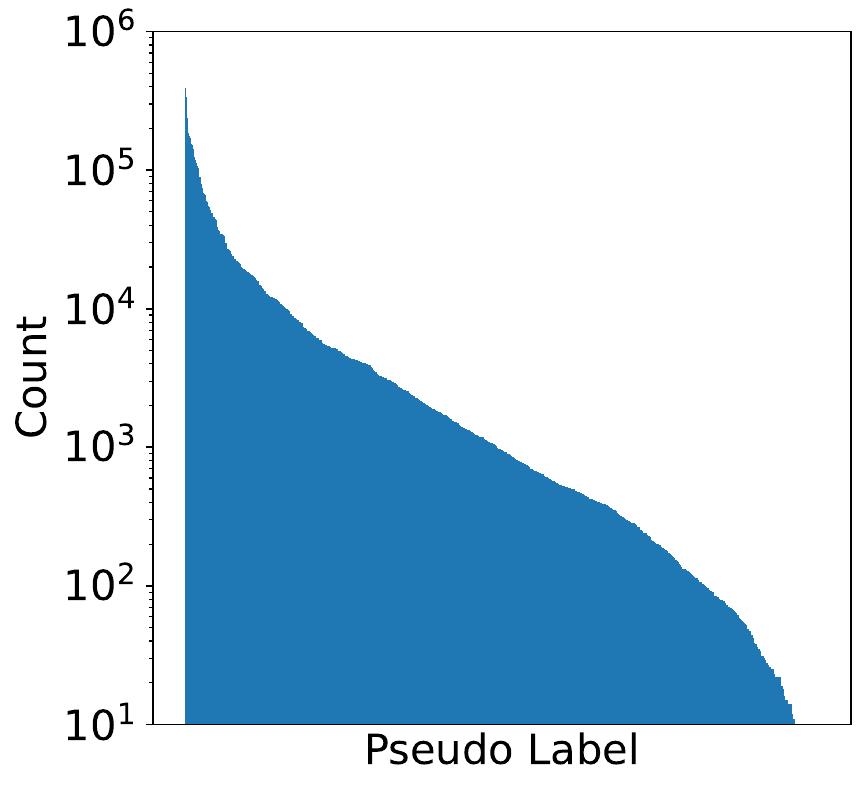}
        \caption{Stage 2}
    \end{subfigure}
    \caption{Distribution of pseudo labels across stages 0, 1, and 2, demonstrating the variation in learned features.}
    \label{fig:pseudo_label_distribution}
\end{figure}

% Further, the distribution of pseudo labels for each stage, depicted in Figure~\ref{fig:pseudo_label_distribution}, showcases a long-tail distribution. This observation is consistent with the characteristic distributions seen in large-scale datasets~\cite{assran2022hidden,tian2021divide}, highlighting the natural variety in the data captured at different stages of MCL.
Additionally, the distribution of pseudo labels for each stage, depicted in \cref{fig:pseudo_label_distribution} (log scale), exhibits a long-tail distribution. This pattern aligns with characteristic distributions observed in large-scale datasets~\cite{assran2022hidden,tian2021divide}, highlighting the natural variety of data captured at different stages of MCL.

%\bibliographystyle{splncs04}
%\bibliography{supplementary}
%\printbibliography[title={Supplementary References}, keyword=supp]

\end{document}